\documentclass{article}
\usepackage{amsmath, amsthm, amssymb}

\usepackage{expl3}  %
\ExplSyntaxOn
  _to_str:N
\ExplSyntaxOff
\usepackage{ifthen}
\usepackage{pgffor}

\usepackage[acronym,smallcaps,nowarn,section,nonumberlist]{glossaries}
\glsdisablehyper  %
\makeglossaries

\setacronymstyle{long-short}

\newacronym{auc}{AUC}{Area Under the Curve}

\newacronym{bert}{BERT}{Bidirectional Encoder Representations from Transformers}
\newacronym{bow}{BOW}{Bag of Words}
\newacronym{boe}{BOE}{Bag of Embeddings}
\newacronym{borep}{BOREP}{Bag of Random Embedding Projections}
\newacronym{bn}{BN}{Batch Normalization}

\newacronym{cdf}{CDF}{Cumulative Distribution Function}
\newacronym{cnn}{CNN}{Convolutional Neural Network}
\newacronym{cka}{CKA}{Centered Kernel Alignment}
\newacronym{cvae}{CVAE}{Conditional Variational Auto-Encoder}

\newacronym{dag}{DAG}{Directed Acyclic Graph}
\newacronym{dgm}{DGM}{Directed Graphical Model}
\newacronym{dnn}{DNN}{Deep Neural Network}

\newacronym{ecfp4}{ECFP4}{Extended Connectivity Fingerprint}
\newacronym{ehr}{EHR}{Electronic Health Record}
\newacronym{esn}{ESN}{Echo State Network}

\newacronym{fft}{FFT}{Fast Fourier Transform}
\newacronym{fs}{FS}{FastSent}
\newacronym{ft}{FT}{Fourier Transform}

\newacronym{gat}{GAT}{Graph Attention Network}
\newacronym{gcn}{GCN}{Graph Convolutional Network}
\newacronym{gdl}{GDL}{Geometric Deep Learning}
\newacronym{ggnn}{GGNN}{Gated Graph Neural Network}
\newacronym{glove}{GloVe}{Global Vectors for Word Representation}
\newacronym{gnn}{GNN}{Graph Neural Network}
\newacronym{gru}{GRU}{Gated Recurrent Unit}

\newacronym{hmm}{HMM}{Hidden Markov Model}

\newacronym{idf}{IDF}{Inverse Document Frequency}
\newacronym{ig}{IG}{Information Gain}
\newacronym{isan}{ISAN}{Input Switched Affine Network}

\newacronym{jsd}{JSD}{Jensen-Shannon Divergence}

\newacronym{kld}{KLD}{Kullback-Leibler Divergence}
\newacronym{knn}{KNN}{$K$-Nearest Neighbour}

\newacronym{lhs}{L.H.S.}{left hand side}
\newacronym{lstm}{LSTM}{Long Short Term Memory Network}
\newacronym{lv}{LV}{latent variable}

\newacronym{map}{MAP}{Maximum a Posteriori}
\newacronym{mc}{MC}{Monte Carlo}
\newacronym{mdl}{MDL}{Minimum Description Length}
\newacronym{ml}{ML}{Machine Learning}
\newacronym{mlp}{MLP}{Multi-Layer Perceptron}
\newacronym{mle}{MLE}{Maximum Likelihood Estimation}
\newacronym{mnist}{MNIST}{Modified National Institute of Standards and Technology}

\newacronym{nlp}{NLP}{Natural Language Processing}
\newacronym{nn}{NN}{Neural Network}
\newacronym{nll}{NLL}{Negative Log Likelihood}
\newacronym{nce}{NCE}{Noise Contrastive Estimation}

\newacronym{ode}{ODE}{Ordinary Differential Equation}
\newacronym{oh}{OH}{One-Hot}

\newacronym{pdf}{PDF}{Probability Density Function}
\newacronym{pgm}{PGM}{Probabilistic Graphical Model}

\newacronym{relu}{ReLU}{Rectified Linear Unit}
\newacronym{rdf}{RDF}{Resource Description Framework}
\newacronym{rgb}{RGB}{Red Green Blue}
\newacronym{rgc}{RGC}{Relational Graph Convolution}
\newacronym{rgat}{RGAT}{Relational Graph Attention Network}
\newacronym{rgcn}{RGCN}{Relational Graph Convolutional Network}
\newacronym{rhs}{R.H.S.}{Right Hand Side}
\newacronym{rnn}{RNN}{Recurrent Neural Network}
\newacronym{roc}{ROC}{Receiver Operating Characteristic}
\newacronym{rv}{RV}{Random Variable}

\newacronym{sgd}{SGD}{Stochastic Gradient Descent}
\newacronym{sg}{SG}{SkipGram}
\newacronym{simclr}{SimCLR}{A Simple framework for Contrastive Learning of visual Representations}
\newacronym{ssl}{SSL}{Self-Supervised Learning}
\newacronym{st}{ST}{SkipThought}
\newacronym{sts}{STS}{Semantic Textual Similarity}
\newacronym{scon}{StochCon}{Stochastic Contrastive Learning}
\newacronym{supcon}{SupCon}{Supervised Contrastive Learning}

\newacronym{termfreq}{TF}{Term Frequency}
\newacronym{tfidf}{TF-IDF}{Term Frequency-Inverse Document Frequency}
\newacronym[plural=TPPs, descriptionplural={Temporal Point Processes}]{tpp}{TPP}{Temporal Point Process}

\newacronym{vae}{VAE}{Variational Auto-Encoder}

\newacronym{wirgat}{WIRGAT}{Within-Relation Graph Attention}
\newacronym{wl}{WL}{Weisfeiler-Lehman}

\usepackage{bm}
\usepackage[T1]{fontenc}  %
\usepackage[type1]{libertine}
\usepackage[libertine]{newtxmath}
\usepackage{mathtools}

\usepackage{tikz}
\usetikzlibrary{positioning,decorations.pathmorphing,bayesnet}

\usepackage{algorithm}
\usepackage{algpseudocode}

\usepackage{cleveref}
\usepackage{comment}
\usepackage{booktabs}

\usepackage{listings}  %
\usepackage{dsfont}  %

\DeclareRobustCommand{\mathup}[1]{\begingroup\changegreek\mathrm{#1}\endgroup}
\DeclareRobustCommand{\mathbfup}[1]{\begingroup\changegreekbf\mathbf{#1}\endgroup}
\DeclareRobustCommand{\mathbit}[1]{\bm{\mathit{#1}}}

\DeclareMathAlphabet{\mathsfit}{\encodingdefault}{\sfdefault}{m}{sl}
\SetMathAlphabet{\mathsfit}{bold}{\encodingdefault}{\sfdefault}{bx}{n}
\newcommand{\tens}[1]{\bm{\mathsfit{#1}}}

\newcommand{\constantvector}{\bm}               %

\newcommand{\constantmatrix}{\bm}               %
\newcommand{\constantmatrixgreek}{\mathbit}

\newcommand{\randomscalar}{\textnormal}         %
\newcommand{\randomscalargreek}{\mathup}

\newcommand{\randomvector}{\mathbf}             %
\newcommand{\randomvectorgreek}{\mathbfup}

\newcommand{\randommatrix}{\mathbf}             %
\newcommand{\randommatrixgreek}{\mathbfup}

\newcommand{\graphstyle}{\mathcal}              %

\newcommand{\tensorstyle}{\tens}                %

\newcommand{\setstyle}{\mathbb}                %

\usepackage[hyphens]{url}

\usepackage{xcolor}

\definecolor{codegreen}{rgb}{0,0.6,0}
\definecolor{codegray}{rgb}{0.5,0.5,0.5}
\definecolor{codepurple}{rgb}{0.58,0,0.82}
\definecolor{backcolour}{rgb}{0.95,0.95,0.92}

\lstdefinestyle{codestyle}{
    backgroundcolor=\color{backcolour},
    commentstyle=\color{codegreen},
    keywordstyle=\color{magenta},
    numberstyle=\tiny\color{codegray},
    stringstyle=\color{codepurple},
    basicstyle=\ttfamily\footnotesize,
    breakatwhitespace=false,
    breaklines=true,
    captionpos=b,
    keepspaces=true,
    numbers=left,
    numbersep=5pt,
    showspaces=false,
    showstringspaces=false,
    showtabs=false,
    tabsize=2
}

\def\alphabet{a,b,c,d,e,f,g,h,i,j,k,l,m,n,o,p,q,r,s,t,u,v,w,x,y,z}
\def\Alphabet{A,B,C,D,E,F,G,H,I,J,K,L,M,M,O,P,Q,R,S,T,U,V,W,X,Y,Z}
\def\greekalphabet{alpha,beta,gamma,delta,epsilon,varepsilon,zeta,eta,theta,vartheta,iota,kappa,varkappa,lambda,mu,nu,xi,pi,varpi,rho,varrho,sigma,varsigma,tau,upsilon,phi,varphi,chi,psi,omega}
\def\GreekAlphabet{Gamma,Delta,Theta,Lambda,Xi,Pi,Sigma,Upsilon,Phi,Psi,Omega}

\makeatletter
\def\changegreek{\@for\next:=\greekalphabet
	\do{\expandafter\let\csname\next\expandafter\endcsname\csname\next up\endcsname}}
\def\changegreekbf{\@for\next:=\greekalphabet
	\do{\expandafter\def\csname\next\expandafter\endcsname\expandafter{%
			\expandafter\bm\expandafter{\csname\next up\endcsname}}}}
\makeatother

\foreach \x in \alphabet {
	\expandafter\xdef\csname v\x\endcsname{\noexpand\ensuremath{\noexpand\constantvector{\x}}}

	\expandafter\xdef\csname ev\x\endcsname{\noexpand\ensuremath{\noexpand\x}}

	\ifthenelse{\equal{\x}{m}}{}{
		\expandafter\xdef\csname r\x\endcsname{\noexpand\ensuremath{\noexpand\randomscalar{\x}}}
	}

	\expandafter\xdef\csname rv\x\endcsname{\noexpand\ensuremath{\noexpand\randomvector{\x}}}
}

\foreach \x in \greekalphabet {
	\expandafter\xdef\csname v\x\endcsname{\noexpand\ensuremath{\noexpand\constantvector{\csname \x\endcsname}}}

	\expandafter\xdef\csname ev\x\endcsname{\noexpand\ensuremath{\noexpand{\csname \x \endcsname}}}

	\expandafter\xdef\csname r\x\endcsname{\noexpand\ensuremath{\noexpand\randomscalargreek{\csname \x\endcsname}}}

	\expandafter\xdef\csname rv\x\endcsname{\noexpand\ensuremath{\noexpand\randomvectorgreek{\csname \x\endcsname}}}
}

\foreach \x in \Alphabet {
	\expandafter\xdef\csname m\x\endcsname{\noexpand\ensuremath{\noexpand\constantmatrix{\x}}}

	\expandafter\xdef\csname em\x\endcsname{\noexpand\ensuremath{\noexpand\x}}

	\expandafter\xdef\csname rm\x\endcsname{\noexpand\ensuremath{\noexpand\randommatrix{\x}}}

	\expandafter\xdef\csname t\x\endcsname{\noexpand\ensuremath{\noexpand\tensorstyle{\x}}}

	\expandafter\xdef\csname g\x\endcsname{\noexpand\ensuremath{\noexpand\graphstyle{\x}}}

	\ifthenelse{\equal{\x}{E}}{}{
		\expandafter\xdef\csname s\x\endcsname{\noexpand\ensuremath{\noexpand\setstyle{\x}}}
	}

}

\foreach \x in \GreekAlphabet {
	\expandafter\xdef\csname m\x\endcsname{\noexpand\ensuremath{\noexpand\constantmatrixgreek{\csname \x\endcsname}}}

	\expandafter\xdef\csname rm\x\endcsname{\noexpand\ensuremath{\noexpand\randommatrixgreek{\csname \x\endcsname}}}
}

\newcommand{\R}{\mathbb{R}}

\newcommand{\x}{\times}

\usepackage[final]{templates/neurips2021/neurips_2021}

\title{Stochastic Contrastive Learning}

\newcommand*\samethanks[1][\value{footnote}]{\footnotemark[#1]}

\author{
  Jason Ramapuram\thanks{Equal contribution.}, \ \ \ \ \ Dan
  Busbridge\samethanks, \ \ \ \ \ Xavier Suau, \ \ \ \ \ Russ Webb \\
  Apple \\
\texttt{\{jramapuram,\ dbusbridge,\ xsuaucuadros,\ rwebb\}@apple.com} \\
}

\begin{document}
\maketitle
\begin{abstract}
	While state-of-the-art contrastive \gls{ssl} models produce results
	competitive with their supervised counterparts, they lack the ability to infer
	latent variables.
	In contrast, prescribed \gls{lv} models enable attributing
	uncertainty, inducing task specific compression, and in general allow for more
	interpretable representations.
	In this work, we introduce \gls{lv}
	approximations to large scale contrastive \gls{ssl} models.
	We demonstrate that
	this addition improves downstream performance (resulting in 96.42\% and
	77.49\% test top-1 fine-tuned performance on CIFAR10 and ImageNet respectively
	with a ResNet50) as well as producing highly compressed representations
	(\emph{588$\x$} reduction) that are useful for interpretability, classification
	and regression downstream tasks.
\end{abstract}

\section{Introduction}
\label{sec:introduction}

Learning meaningful representations without human domain knowledge has been a
long-standing goal of machine learning.
Recent work in large scale
\gls{ssl}
\citep{DBLP:conf/icml/ChenK0H20,DBLP:conf/nips/ChenKSNH20,DBLP:conf/nips/AlayracRSARFSDZ20,DBLP:conf/nips/GrillSATRBDPGAP20,DBLP:conf/nips/CaronMMGBJ20,DBLP:journals/corr/abs-2103-03230,DBLP:journals/corr/abs-2104-14294}
has advanced this pursuit and narrowed the gap against fully supervised models,
all the while relaxing the use of potentially biased human labels.
And yet, the \gls{ssl} methods and toolkits lack a method to add interpretable, prescribed distributions into the representation learning process.
In this work, we address
this shortcoming through the introduction of Bernoulli and Isotropic-Gaussian latent
variables into the SimCLR \citep{DBLP:conf/icml/ChenK0H20} contrastive learning
framework.

The use of Bernoulli latent variables enables extracting meaningful discrete
representations of image data, providing a natural means of data dependent
compression\footnote{The
	map $\R^{224 \times 224 \times 3} \mapsto \mathbb{Z}_{2}^{2048}$ implies a
	$\frac{1}{588}\times$ compression given 8 bits/input pixel and 1 bit/binary
	output.} that is useful for downstream tasks such as classification and
regression. Interestingly, we find that the use of discrete latent variables
improves downstream performance when fully finetuning the representation
learning backbone, outperforming SimCLR \citep{DBLP:conf/icml/ChenK0H20} on
CIFAR10 and ImageNet1000 \citep{DBLP:conf/cvpr/DengDSLL009}.

\section{Background}
\label{sec:background}

In this work we focus on large scale contrastive learning, where we optimize
the InfoNCE objective
\citep{DBLP:journals/corr/abs-1807-03748,DBLP:conf/icml/ChenK0H20}. InfoNCE
generalizes
\gls{nce}
by using variates from the empirical data distribution,
$\{\vx_{i}, \vx_{j}\} \sim p(\vx)$, mapping them through networks,
$g_{\vtheta}$ and $f_{\vtheta}$, to a representation
$\vv = (g_{\vtheta} \circ f_{\vtheta} )(\vx)$. $f_{\vtheta}$ is typically
referred to as the backbone and $g_{\vtheta}$ as the InfoNCE head.
While NCE samples negative variates from a naive prior, $\vv \sim p(\vv)$,
InfoNCE uses true variates in a multi-sample un-normalized bound:
\citep{DBLP:conf/icml/PooleOOAT19}
\begin{align}
	\mathcal{L}_\text{InfoNCE}^{(i,j)} & = - \log\frac{\exp(\text{sim}({\vv}_i, {\vv}_j) / \tau)}{\sum_{k=1}^{2N} {\mathds{1}}_{[k \neq i]} \exp(\text{sim}({\vv}_i, {\vv}_k) / \tau)}.
	\label{eqn:infonce}
\end{align}
The similarity operator ($\text{sim}$) from Equation \ref{eqn:infonce} typically
is modeled with a cosine-similarity on the representation feature vectors
$\{\vv_{i}, \vv_{j}\}$, with a controllable temperature hyper-parameter $\tau$.

\section{\gls{scon}}
\label{sec:model}

\newcommand{\DBackbone}{D_{\textnormal{Backbone}}}
\newcommand{\DLatent}{D_{\textnormal{Latent}}}

\begin{figure}[h]
	\centering
	\newcommand{\xsep}{1.5}
\newcommand{\ysep}{1.1}
\newcommand{\nodesize}{23pt}
\newcommand{\platesep}{0.2cm}

\scalebox{0.85}{\parbox{1.0\linewidth}{%
\begin{center}
\begin{tikzpicture}
	\node[draw, circle, minimum size=\nodesize] at (0,0) (x) {$\vx$};
	\node[draw, circle, minimum size=\nodesize] at (\xsep,\ysep) (x1) {$\tilde{\vx}$};
	\node[draw, circle, minimum size=\nodesize] at (\xsep,-\ysep) (x2) {$\tilde{\vx}^{\prime}$};
	\node[draw, circle, minimum size=\nodesize] at (5*\xsep,\ysep) (h1) {$\vh$};
	\node[draw, circle, minimum size=\nodesize] at (2*\xsep,-\ysep) (h2) {$\vh^\prime$};
	\node[draw, circle, minimum size=\nodesize] at (6*\xsep,\ysep) (v1) {$\vv$};
	\node[draw, circle, minimum size=\nodesize] at (6*\xsep,-\ysep) (v2) {$\vv^{\prime}$};

	\node[draw, circle, minimum size=\nodesize] at (3*\xsep,-\ysep) (phi) {$\vphi^{\prime}$};
	\node[draw, circle, minimum size=\nodesize, fill=black!15] at (4*\xsep,-\ysep) (z2) {$\rvz^{\prime}$};
	\node[draw, circle, minimum size=\nodesize] at (5*\xsep,-\ysep) (h2p) {$\vh^{\prime\prime}$};

	\plate [inner sep=\platesep,xshift=0.0cm,yshift=0.0cm] {plate1} {(z2)(v2)} {$K$}; %
	\plate [inner sep=\platesep,xshift=0.0cm,yshift=0.0cm] {plate2} {(x)(v1)(v2)(plate1)} {$N$}; %

	\path[->]
	(x)  edge [>=latex] node[above,rotate=36] {$t\sim\sT$} (x1)
	(x)  edge [>=latex] node[below,rotate=-36] {$t^{\prime}\sim\sT$} (x2)

	(x1)  edge [>=latex] node[above,rotate=0] {$f$} (h1)
	(x2)  edge [>=latex] node[below,rotate=0] {$f$} (h2)

	(h1)  edge [>=latex] node[above,rotate=0] {$g$} (v1)

	(h2)  edge [>=latex] node[below,rotate=0] {$\pi$} (phi)
	(phi)  edge [>=latex] node[below,rotate=0] {} (z2)
	(z2)  edge [>=latex] node[below,rotate=0] {$\rho$} (h2p)
	(h2p)  edge [>=latex] node[below,rotate=0] {$g$} (v2)

	(v1)  edge [latex-latex, black!40] node[left,rotate=0] {Maximize sim} (v2)
	;
  \end{tikzpicture}
  \end{center}
}}
	\caption{
	\gls{scon} without multi-crop \citep{DBLP:conf/nips/CaronMMGBJ20} using plate notation.
	Following~\citet{DBLP:conf/icml/ChenK0H20}, two data augmentation operators $t,t^{\prime}$ are sampled from the augmentation family $\sT$, producing views $\tilde \vx, \tilde \vx^{\prime}$.
	A backbone network $f$ is applied to each example, producing
	$\vh,\vh^{\prime}\in\R^{\DBackbone}$
	(we include spatial pooling in our definion of $f$).
	On the bottom branch, the $\vh^{\prime}$ specifies the natural parameters of a distribution $q(\rvz;\pi(\vh))$, via an encoder
	$\pi:\R^{\DBackbone}\rightarrow \R^{\DLatent}$.
	Variates $\vz^{\prime}\sim q$ are decoded by
	$\rho:\R^{\DLatent}\rightarrow \R^{\DBackbone}$
	and compared using the standard $\mathcal{L}_{{\textnormal{InfoNCE}}}$ (Equation \ref{eqn:infonce}) using the head $g$.
	This work's contribution is the stochastic mapping $\vphi^{\prime}\mapsto \rvz^{\prime} \mapsto \vh^{\prime\prime}$. Gray circles denote random variables.
	Bottom branch negative variates are reparameterized in the same manner.
	}
	\label{fig:architecture}
\end{figure}
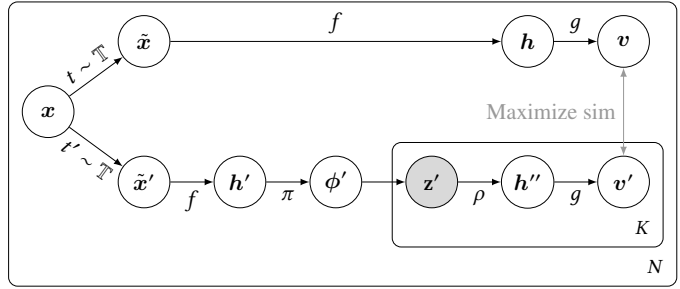

\begin{algorithm}[H]
	\caption{\gls{scon}}\label{alg:stochastic_contrastive_learning}
	\begin{algorithmic}
		\Require Data: $\vx \sim p(\vx), t \sim \sT(\vx)$
		\Require Models: $f_{\vtheta}: \text{backbone}, g_{\vtheta}: \text{head}, \{\pi_{\vtheta}, \rho_{\vtheta}\}:\text{projectors}$
		\While{not converged}
		\State $\{\bm{\hat{x}}, \bm{\hat{x}}^{\prime}\} = \{t \circ \vx, t^{\prime} \circ \vx\}$
		\algorithmiccomment{Augment input with $\{t, t^{\prime}\}$}
		\State $\{\vh, \vh^{\prime}\} = \{f_{\vtheta}(\bm{\hat{x}}), f_{\vtheta}(\bm{\hat{x}}^{\prime})\}$
		\algorithmiccomment{Produce representations}
		\State $\vphi^{\prime} = \pi_{\vtheta}(\vh^{\prime})$
		\algorithmiccomment{(optional) Bottleneck projection}
		\State $\rvz^{\prime} \sim q_{\vtheta}(\rvz|\vx)$
		\algorithmiccomment{Pathwise differentiable \citep{DBLP:journals/jmlr/MohamedRFM20} latent variable.}
		\State $\vh^{\prime \prime} = \rho_{\vtheta}(\rvz^{\prime})$
		\algorithmiccomment{(optional) Bottleneck upsampler}
		\State $\{\vv, \vv^{\prime}\} = \{g_{\vtheta}(\vh), g_{\vtheta}(\vh^{\prime \prime})\}$
		\algorithmiccomment{InfoNCE projection}
		\State $\min_{\vtheta} \mathcal{L}_{\text{InfoNCE}}(\vv, \vv^{\prime})$
		\EndWhile
	\end{algorithmic}
      \end{algorithm}
We describe our model in Algorithm \ref{alg:stochastic_contrastive_learning} and
Figure \ref{fig:architecture}.
We modify
SimCLR \citep{DBLP:conf/icml/ChenK0H20} by forcing bottom branch variates
through a pathwise differentiable
\citep{DBLP:journals/jmlr/MohamedRFM20} distribution,
$\rvz^{\prime} \sim q_{\vtheta}(\rvz|\vx)$.
Importantly,
$\rvz^{\prime}$ can be optionally
projected to a lower dimensional space,
$|\rvz^{\prime}| << |\vh|$, through  linear projection layers,
$\{\pi_{\vtheta}, \rho_{\vtheta}\}$.
Upon ablation, we observe minimal performance degradation when projecting one
branch of the SimCLR model through a low dimensional distribution, with the advantage of having more interpretable features
(Section \ref{sec:ablations}).

In this work, we explore the isotropic-Gaussian
\citep{DBLP:journals/corr/KingmaW13} and
Gumbel-Bernoulli \citep{DBLP:conf/iclr/JangGP17,DBLP:conf/iclr/MaddisonMT17} distributions.
We apply the
differentiable distribution on the output of the backbone model $f_{\vtheta}$,
given an optional bottleneck projection $\pi_{\vtheta}$.

\section{Experiments}
\label{sec:experiments}

\paragraph{Training details} Following \citet{DBLP:conf/icml/ChenK0H20}, all models train with a batch size of
4096, the LARS optimizer \citep{DBLP:conf/aaai/HuoGH21} with linear warmup
\citep{DBLP:journals/corr/GoyalDGNWKTJH17} and a single cycle cosine annealed
learning rate schedule
\citep{DBLP:journals/corr/GoyalDGNWKTJH17,DBLP:journals/corr/abs-1708-07120}. We
use DINO augmentations \citep{DBLP:journals/corr/abs-2104-14294} (2-global views
+ 8-local views \citep{DBLP:conf/nips/CaronMMGBJ20}) for all SimCLR variants.
For the Gumbel-Bernouilli distribution, the temperature is annealed from
$1.0 \rightarrow 0.1$ using a single cycle cosine schedule during training.
Finetuning procedure is described in \Cref{sec:finetuning}.

Model performance for linear-probes on a non-updated (\emph{Frozen}) and fine-tuned (\emph{Fine-Tuned}) backbone is given in \Cref{table:standard_metrics}.
We observe that \gls{scon} (\emph{Fine-Tuned})
outperforms an equally tuned SimCLR model,
as well as a supervised model with the same ResNet50 and ResNet200 \citep{DBLP:conf/cvpr/HeZRS16} architectures,
while the \emph{Frozen} probe is competitive.
We validate that this performance difference does not arise purely from the Gumbel-Bernoulli through ablations presented in
\Cref{sec:stochcon_ft_and_supervised_bernoulli}.

\begin{table}[H]
	\caption{Summary test top-1\% for CIFAR10 and ImageNet1000.}
	\label{table:standard_metrics}
	\centering
	\begin{tabular}{lccccccc}
		\toprule
		                   & \multicolumn{2}{c}{CIFAR10-ResNet50} & \multicolumn{2}{c}{ImageNet-ResNet50} & \multicolumn{2}{c}{ImageNet-ResNet200} &                                                  \\
		\cmidrule(r){2-3}
		\cmidrule(r){4-5}
		\cmidrule(r){6-7}
		Model              & Fine-Tuned                           & Frozen                                & Fine-Tuned                             & Frozen         & Fine-Tuned     & Frozen         \\
		\midrule
		StochCon Bern      & \textbf{96.42}                       & 91.96                                 & \textbf{77.49}                         & 67.00          & \textbf{80.24} & 64.25          \\
		StochCon Iso-Gauss & 96.08                                & \textbf{92.40}                        & --                                     & --             & --             & --             \\
		Supervised         & 95.00                                & --                                    & 76.13                                  & --             & 78.34          & --             \\
		SimCLR             & 94.35                                & 91.67                                 & 76.37                                  & \textbf{71.34} & 79.82          & \textbf{73.52} \\
		\bottomrule
	\end{tabular}
\end{table}

\subsection{Ablations}
\label{sec:ablations}

\begin{figure}[H]
      \begin{center}
  \scalebox{0.85}{\parbox{1.0\linewidth}{%
\begin{minipage}{0.49\textwidth}
  \begin{center} \centering
\includegraphics[width=\linewidth]{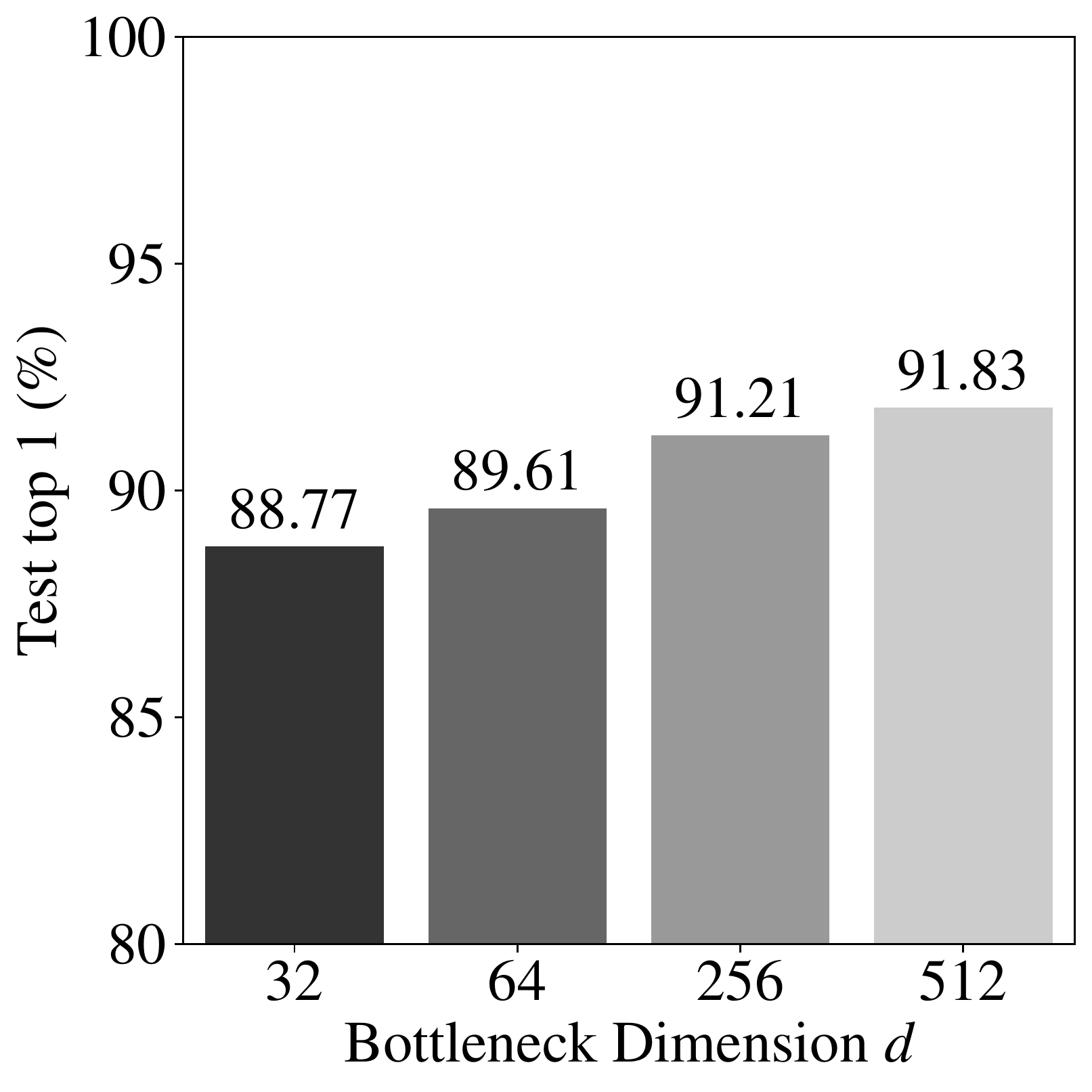}
  \end{center}
\end{minipage}%
\begin{minipage}{0.49\textwidth}
  \begin{center} \centering
\includegraphics[width=\linewidth]{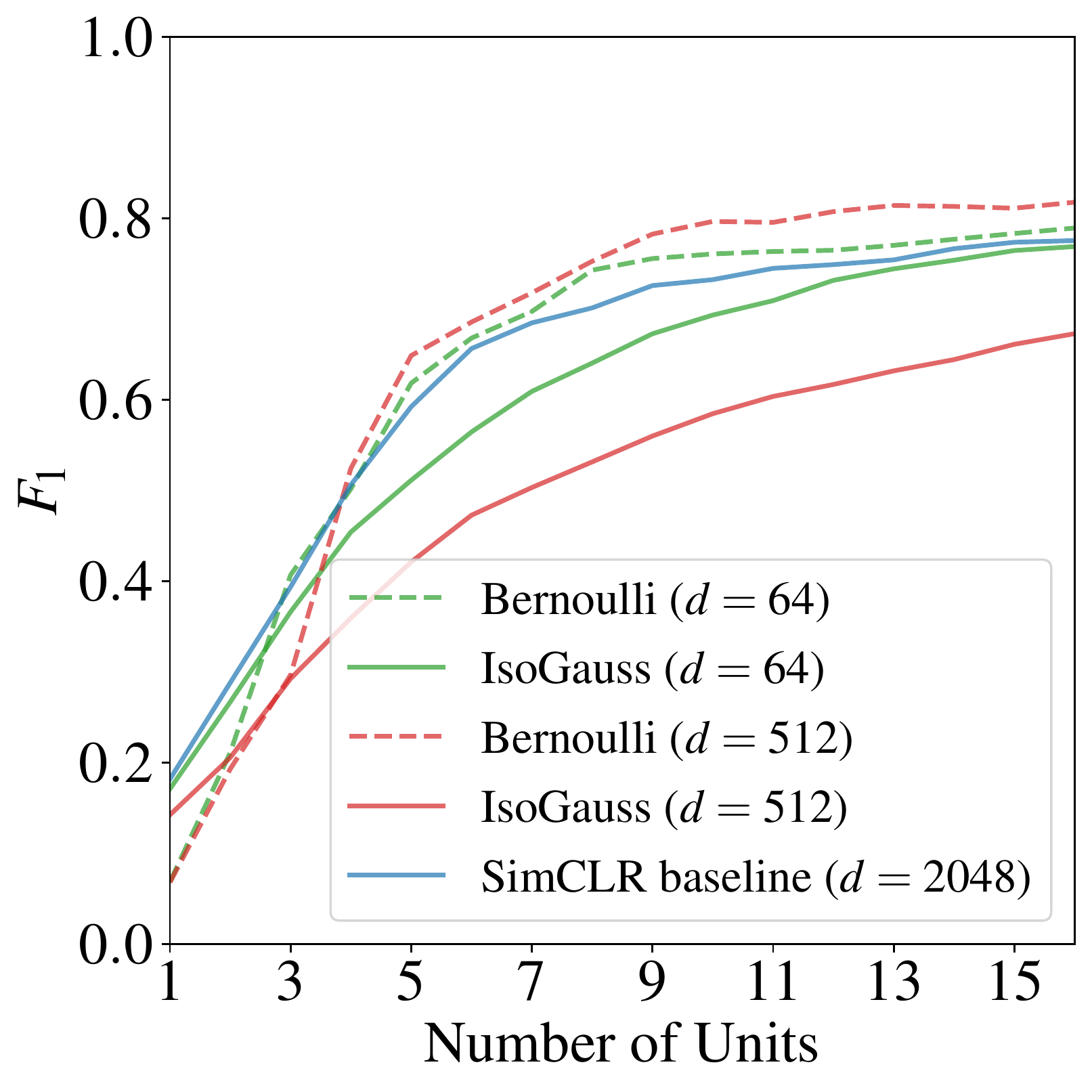}
  \end{center}
\end{minipage}
}}
\end{center}
\vskip -0.1in
\caption{\textbf{Left}: Ablation over bottleneck latent
dimensions for Bernoulli-\gls{scon}. \textbf{Right}: Mean $F_{1}$ performance across all classes on
CIFAR10 for a multi-class Random Forest, varying the number of feature units.
Units are identified on the training set using Random Forest feature importance
under 5-fold stratified sampling. Performance is the mean over the held-out test
sets.
%We see that at each dimensionality, the best-performing models are those specifically trained with the lower-dimensional bottleneck.
}\label{fig:number_of_units}
\end{figure}

To evaluate the benefits of our \gls{scon} framework, we propose a
series of ablations. In Figure \ref{fig:number_of_units}-\emph{Left}, we train \gls{scon} Bernoulli models with varying bottleneck $\vphi^{\prime}$ dimensions and
present the top-1 \emph{Frozen} performance of each model. Results show that \gls{scon}
is robust to variadic sized latents. We believe this robustness is due to the model learning to compare a full $\vh \in \R^{2048}$ dimensional vector to an upsampled low dimensional latent, $\vz^{\prime} \in \R^{D}, D << 2048$. %

In Figure \ref{fig:number_of_units}-\emph{Right}, we analyze the mean $F_{1}$
performance for a multi-class Random Forest evaluated by varying the number of
feature units. Surprisingly, we find that to accurately classify CIFAR10, the
\gls{scon} model with a 64 dimensional latent Gumbel-Bernoulli only requires 11 binary feature units.
We also observe that performance decreases for the Isotropic-Gaussian as we increase the latent dimensionality, holding the number of Random Forest units constant.
Note that this does not happen in the Gumbel-Bernoulli case, as this variable does form distributed representations in the way an Isotropic-Gaussian does.

\vspace{-0.1in}
\vspace{-0.1in}

\subsection{Isotropic-Gaussian \gls{scon} and variance collapse}
\label{sec:iso_gauss_var_collapse}

Since \gls{scon} does not constrain the latent variable distribution, we
observed that in the case of a learned variance where $\rvz^{\prime} \sim \mathcal{N}\big( \vmu(\pi(\vh^{\prime})), \vsigma^{2}(\pi(\vh^{\prime})) \big)$,
the learned variance, $\vsigma^{2}(\pi(\vh^{\prime}))$ would
trivially collapse to 0.
To work around this and provide meaningful
uncertainties, we force the network to learn to estimate variances of the
opposing set of views, so that $\vz^{\prime} \sim \mathcal{N}\big( \vmu(\pi(\vh^{\prime})), \vsigma^{2}(\pi(\vh)) \big)$.
We validate this below in Figure \ref{fig:iso_gauss_collapse} and find that as the bottleneck dimension reduces, the model learns to rely more on the available stochasticity.
\begin{figure}[H]%
\begin{center}
  \scalebox{0.85}{\parbox{1.0\linewidth}{%
\begin{minipage}{0.49\textwidth}
  \begin{center} \centering
\includegraphics[width=\linewidth]{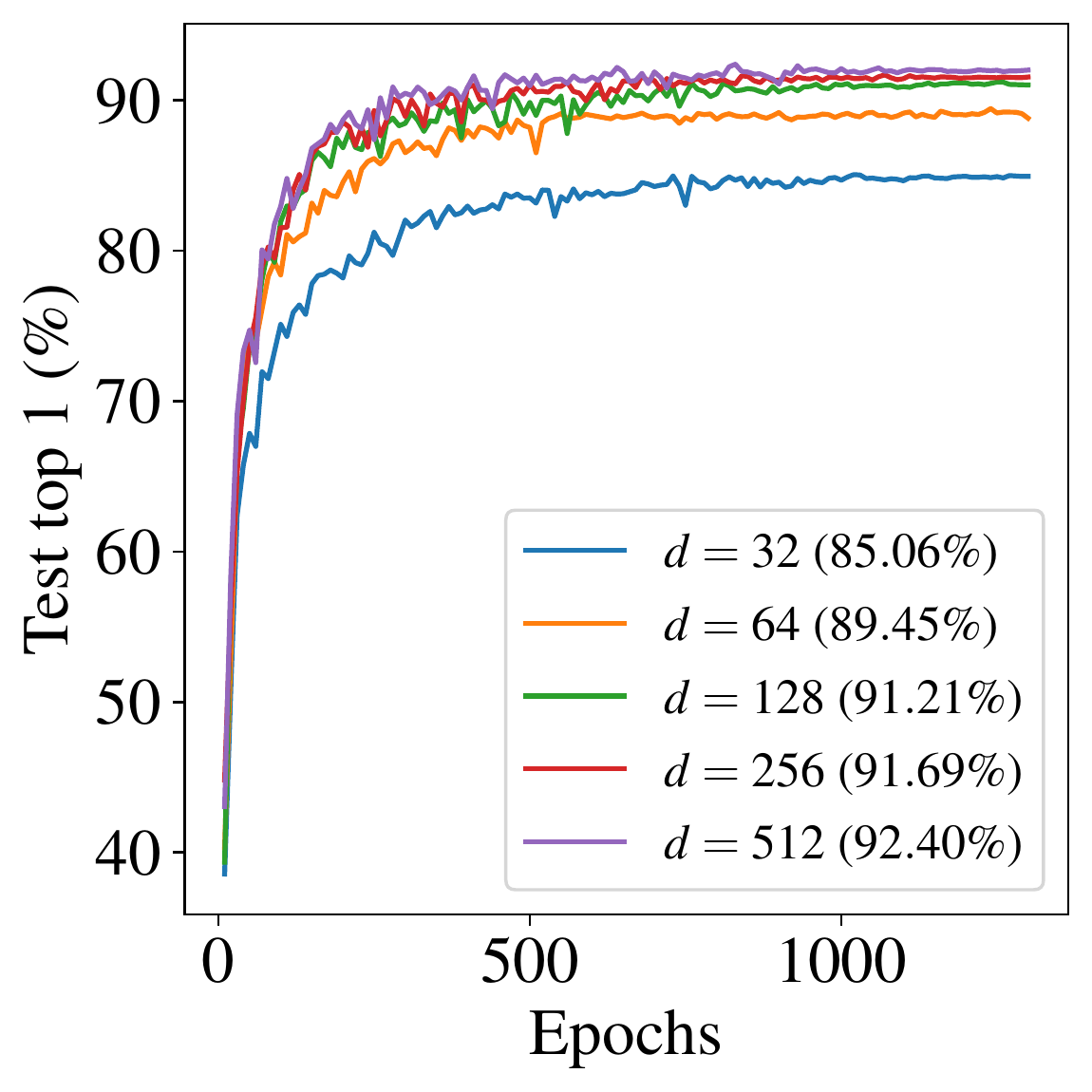}
  \end{center}
\end{minipage}%
\begin{minipage}{0.49\textwidth}
  \begin{center} \centering
\includegraphics[width=\linewidth]{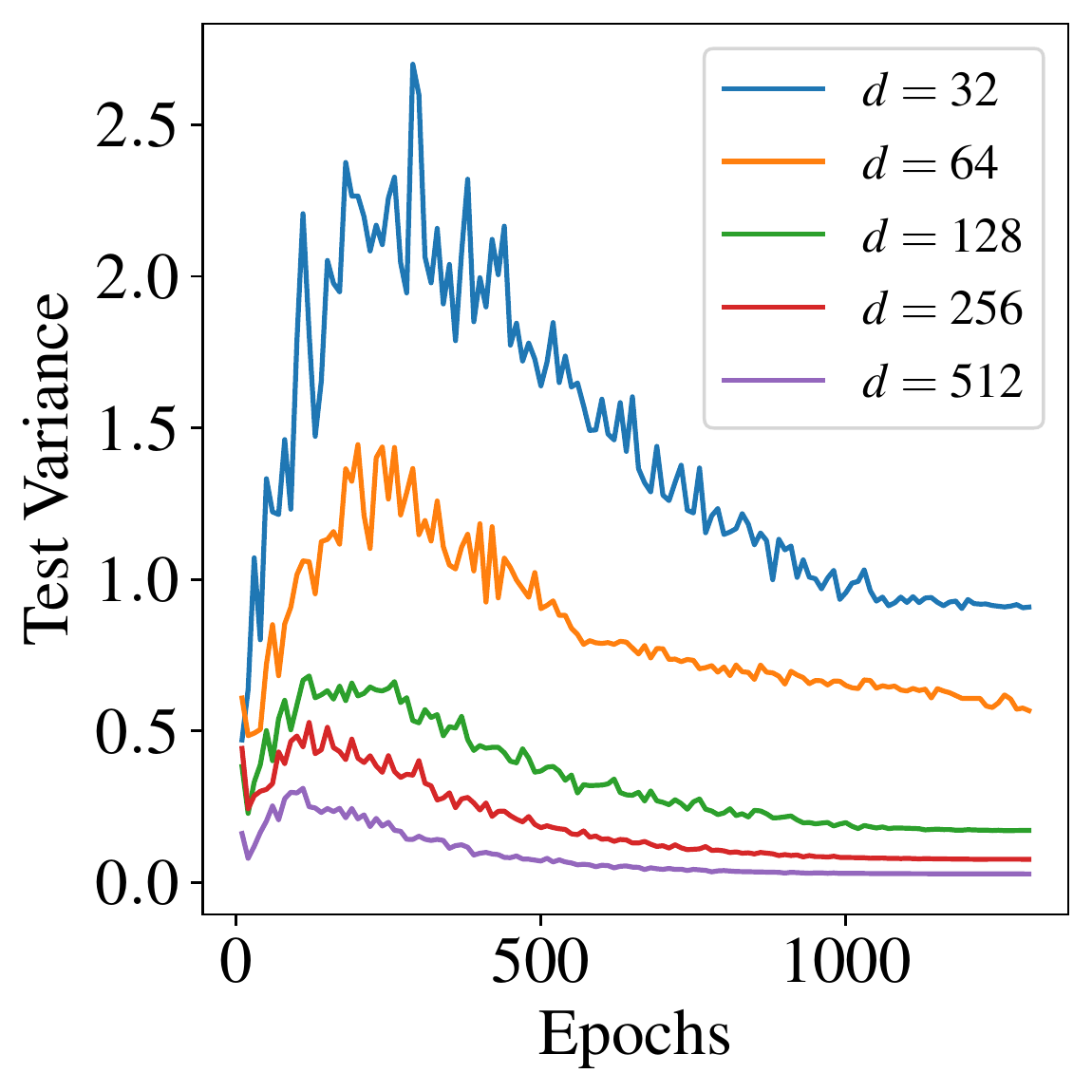}
  \end{center}
\end{minipage}
}}
\end{center}
\caption{CIFAR10 test top-1 (\textbf{Left}) and test aggregate variance, $\mathrm{Var}_{x}[q_{\vtheta}(\vz|\vx)]$, (\textbf{Right}) for varying isotropic-Gaussian bottleneck dimensions.}\label{fig:iso_gauss_collapse}
\vskip -0.1in
\end{figure}

\subsection{Countable metrics for Bernoulli-\gls{scon}}
\label{sec:analysis}

\begin{figure}[H]
\begin{center}
  \scalebox{0.9}{\parbox{1.0\linewidth}{%
\begin{minipage}{0.49\textwidth}
  \begin{center} \centering
\includegraphics[width=\linewidth]{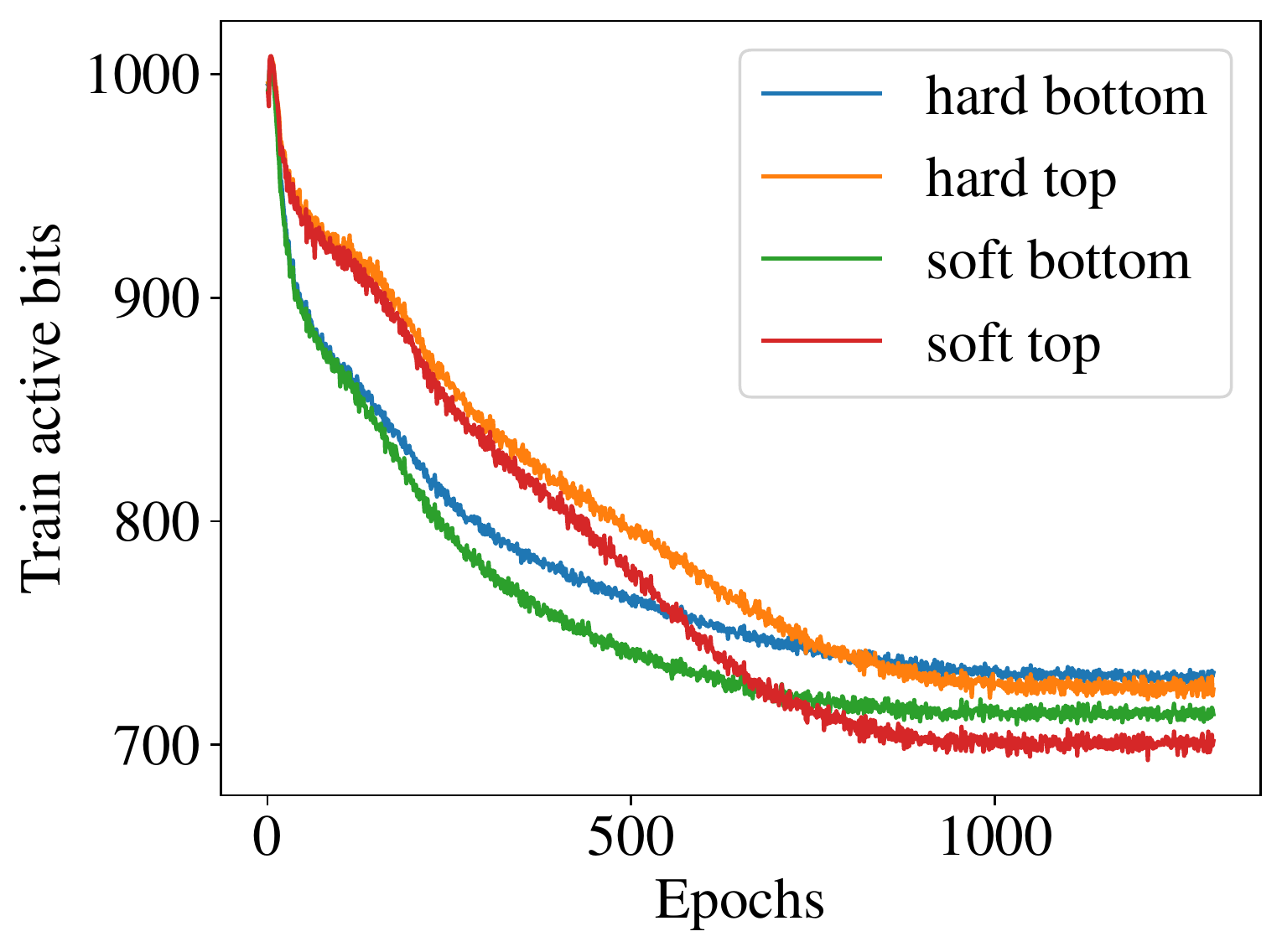}
  \end{center}
\end{minipage}%
\begin{minipage}{0.49\textwidth}
  \begin{center} \centering
\includegraphics[width=\linewidth]{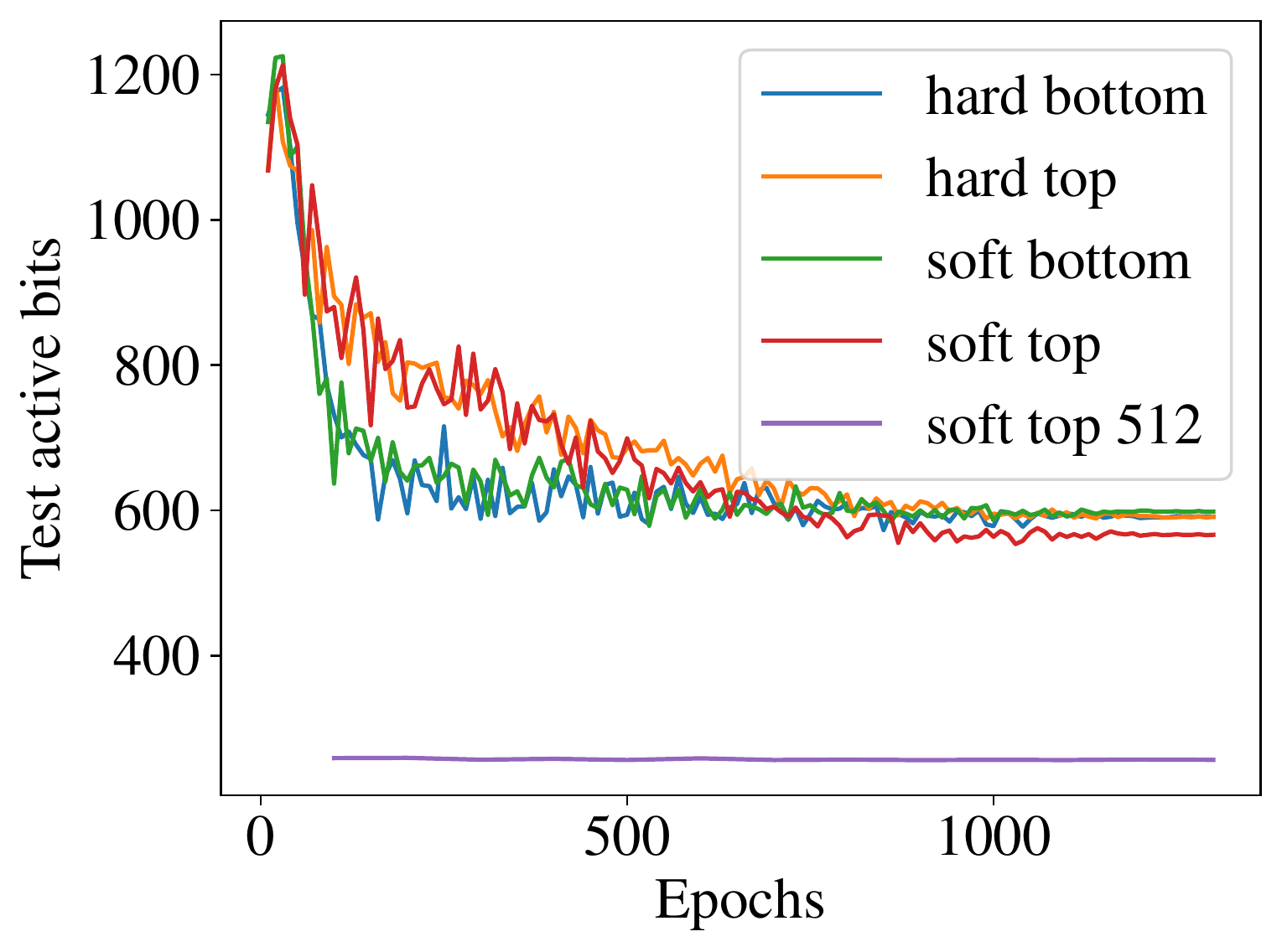}
  \end{center}
\end{minipage}
}}
\end{center}
\vskip -0.1in
\caption{CIFAR10 average bit counts, aggregated across training (\textbf{Left}) and test
(\textbf{Right}) datasets.}\label{fig:active_bits}
\end{figure}

Since \gls{scon} works with discrete representation vectors, it enables analysis
through countable metrics. We present the average count of representation bits
for the $\R^{2048}$ dimensional \gls{scon} Bernoulli model in Figure
\ref{fig:active_bits}. We ablate four different variants: \{\emph{hard bottom},
\emph{hard top}, \emph{soft bottom}, \emph{soft top}\}. The difference between
these variants is where the distribution is applied: the \emph{top-*} models
apply the reparameterization on the global image views
\citep{DBLP:conf/nips/CaronMMGBJ20}, while the \emph{bottom-*} models apply them
on the local views \citep{DBLP:conf/nips/CaronMMGBJ20}. The \emph{hard-*} models
use a differentiable mechanism to always feed-forward discrete variates (see
Appendix Section \ref{sec:discretized_gumbel_bernoulli_variates}), while the
\emph{soft-*} models use the standard variates extracted from the
Gumbel-Bernoulli distribution.

At the beginning of training we observe that the average number of active bits
is approximately half of the available $\R^{2048}$, but as
training progresses this quantity decreases.
Note that this does not imply that
the model does not use the zero valued bits, but rather provides an alternative
method to analyze model performance. For reference in Figure
\ref{fig:active_bits}-\emph{Right} we also include the test active bits for the
512 dimensional bottleneck model (\emph{soft-top 512}). We observe that when the model is capacity restricted it uses all available
Bernoulli latents (50\% of its representation for zeros, 50\% for ones).

\vspace{-0.1in}
\section{Conclusion}
\label{sec:conclusion}

\vspace{-0.1in}
In this work, we present a novel formulation that enables the use of latent
variables in large scale contrastive self-supervised models. We demonstrate that
in addition to improving downstream performance these models reveal that in
practice competitive discriminative performance on CIFAR10 can be achieved with
as few as 11 bits (Figure \ref{fig:number_of_units}). Future work will explore
further latent variable models such as the pathwise Beta distribution
\citep{DBLP:conf/nips/FigurnovMM18} and non-parametric latents such as
normalizing flows \citep{DBLP:conf/icml/RezendeM15}.

\section*{Acknowledgments}

The authors would like to thank the following people for their help throughout
the process of writing this paper, in alphabetical order: Barry-John Theobald,
Katherine Metcalf, Luca Zappella and Miguel Sarabia del Castillo. Additionally,
we thank Andrea Klein, Cindy Liu, Guihao Liang, Guillaume Seguin, Li Li, Okan
Akalin, and the wider Apple infrastructure team for assistance with developing
scalable, fault tolerant code.

\bibliography{libraries/distributions,libraries/selfsup,libraries/celeba,libraries/models,libraries/other,user/library}
\bibliographystyle{templates/iclr2021/iclr2021_conference}

\clearpage
\appendix
\section{Appendix}
\label{sec:appendix}
\vspace{-0.1in}

\subsection{Discrete Gumbel-Bernoulli variates}
\label{sec:discretized_gumbel_bernoulli_variates}

The Gumbel-Bernoulli distribution in its naive form returns non discretized
variates when the temperature, $\tau$, is high. However, a well known trick to
extract proper discrete variates is summarized in the pytorch code below.

\begin{lstlisting}[language=Python, caption=Hard Gumbel-Bernoulli variates.]
import torch

def compute_hard(relaxed: torch.Tensor) -> torch.Tensor:
    """Produce a hard version of relaxed as a differentiable tensor.

    :param relaxed: the relaxed estimate
    :returns: hard bernoulli with 0s and 1s
    :rtype: torch.Tensor

    """
    hard = relaxed.clone()
    hard[relaxed < 0.5] = 0.0
    hard[relaxed >= 0.5] = 1.0
    hard_diff = hard - relaxed  # sub the relaxed tensor backprop path
    return hard_diff.detach() + relaxed  # add back to keep bp path
\end{lstlisting}

\vspace{-0.1in}
\subsection{SimCLR Finetuning and Supervised Bernoulli}
\label{sec:stochcon_ft_and_supervised_bernoulli}
\vspace{-0.1in}

\begin{figure}[H]
	\begin{minipage}{0.49\textwidth}
		\begin{center} \centering
			\includegraphics[width=\linewidth]{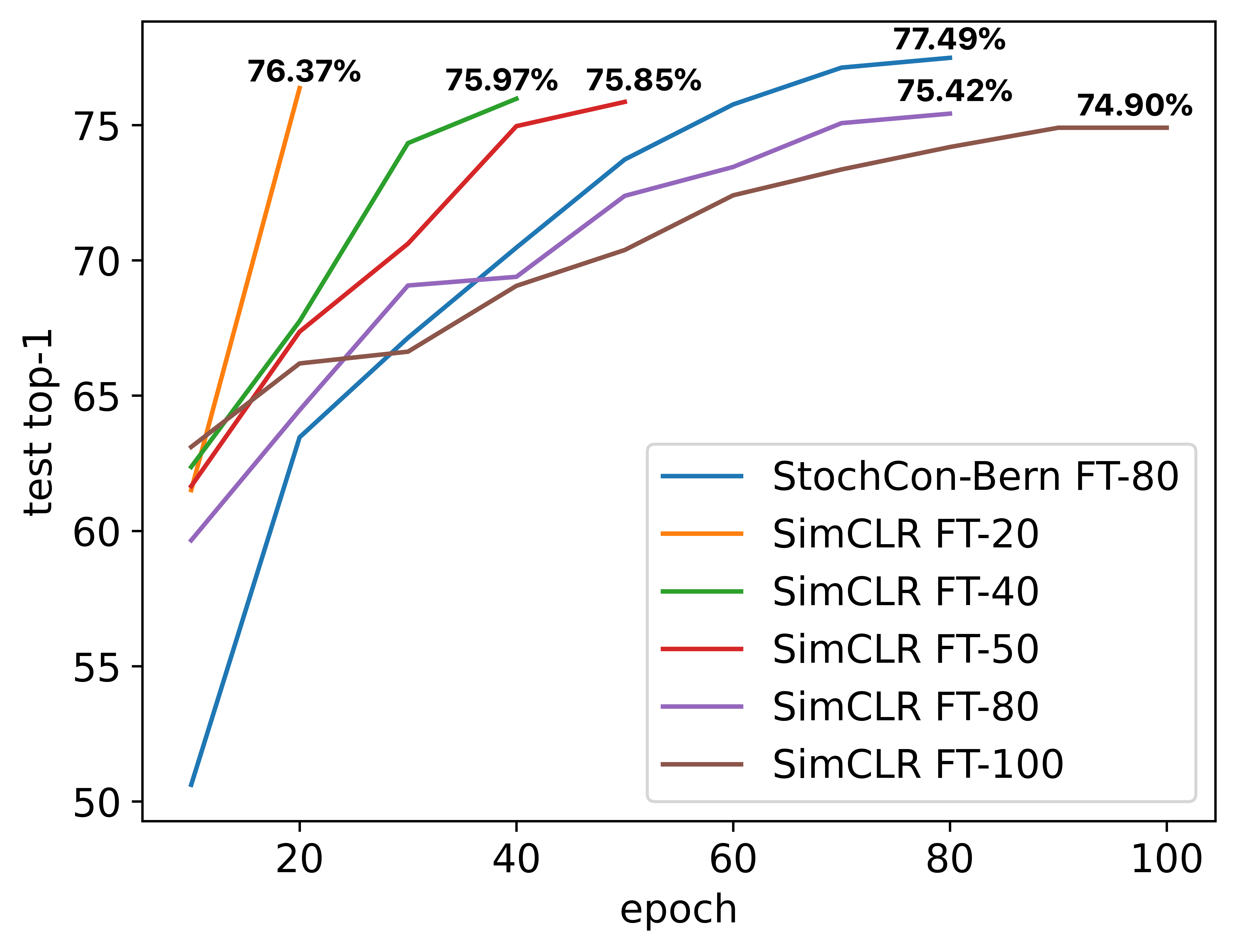}
		\end{center}
	\end{minipage}%
	\begin{minipage}{0.49\textwidth}
		\begin{center} \centering
			\includegraphics[width=\linewidth]{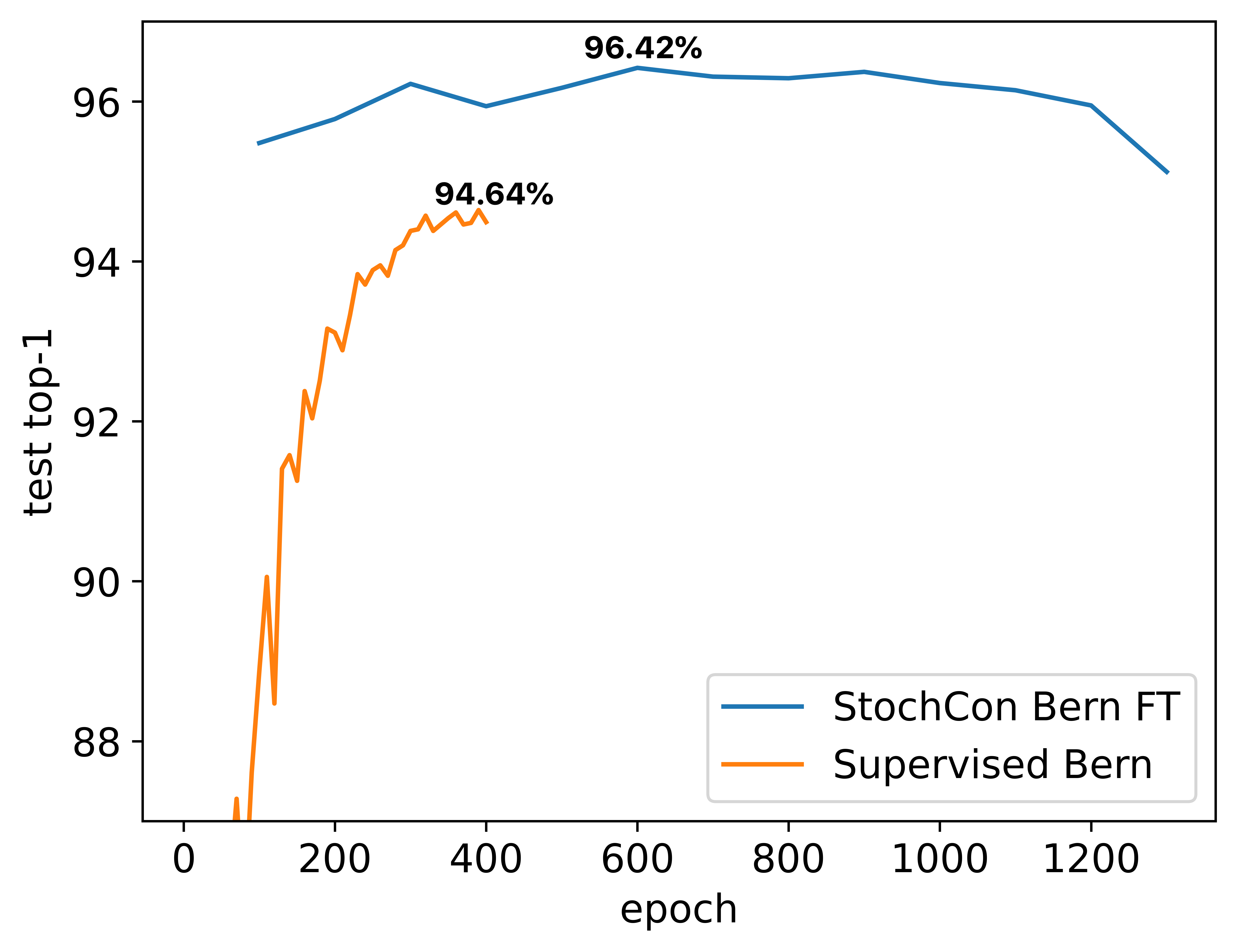}
		\end{center}
	\end{minipage}
	\caption{\textbf{Left}: Test top-1 performance of multiple trials of SimCLR
		finetuning on ImageNet. \textbf{Right}: Adding a Gumbel Bernoulli to the
		representation layer ( with $p=0.5$ ) of a standard ResNet-50 and training
		end-to-end on CIFAR10.}\label{fig:finetune_and_supbern}
\end{figure}

To validate that the performance difference in Table \ref{table:standard_metrics} was not purely from the finetuning process we perform two experiments:

\begin{enumerate}
	\item \textbf{ImageNet finetuning:} In Figure
	      \ref{fig:finetune_and_supbern}-\emph{Left} we finetune multiple SimCLR models
	      over various epoch intervals: \{20, 40, 60, 80, 100\} and observe that SimCLR
	      does not exceed the reported 76.37\% top-1 reported in Table
	      \ref{table:standard_metrics} for ImageNet.

	\item \textbf{CIFAR10 Bernoulli:} In Figure
	      \ref{fig:finetune_and_supbern}-\emph{Right} we add a Gumbel Bernoulli layer to
	      the final layer of a standard ResNet50 model (after spatial pooling) and train
	      the model in a standard supervised setting, dropping out the Gumbel-Bernoulli
	      layer with $p=0.5$. The dropout of the layer functions as a proxy to the branch
	      mechanism used in \gls{scon}. We present the best performing model
	      \footnote{Note that supervised learning typically does not benefit from longer
		      training durations without the use of strong augmentations
		      \citep{DBLP:journals/corr/abs-1805-09501,DBLP:conf/cvpr/CubukZSL20}.}) and note
	      that \gls{scon} outperforms the baseline by 1.78\%.
\end{enumerate}

\vspace{-0.1in}
\subsection{Finetuning procedure}
\label{sec:finetuning}
\vspace{-0.1in}

To finetune \gls{scon}, we retain the pre-trained backbone and latent variable
distribution, and finetune with Adam \citep{DBLP:journals/corr/KingmaB14}. The
\emph{Finetuned} model updates the parameters of the entire network (including
the backbone and newly attached linear head), while the \emph{Frozen} model only
updates the added linear projection head. We use a learning rate of 3e-4,
coupled with a simple step scheduler that scales the learning rate by 0.1 at
80\% of training. All our models (including baselines) are trained for various
epoch ranges using standard ImageNet augmentations (random flip, random-resized
crop), and we report the best performing model in Table
\ref{table:standard_metrics}.

For the Bernoulli-\gls{scon} model, we set the temperature to 0.1 for the entire
finetuning process, while the Isotropic-Gaussian distribution only uses the mean
(similar to Variational Autoencoders \citep{DBLP:journals/corr/KingmaW13} at
inference time). We suspect that the performance of the \emph{Frozen}
Bernoulli-\gls{scon} will match the \emph{Frozen} Isotropic-Gaussian-\gls{scon}
model with a properly tuned Gumbel-Bernoulli temperature schedule
\citep{DBLP:conf/iclr/JangGP17,DBLP:conf/iclr/MaddisonMT17}, but leave this for
future work.

\end{document}